\documentclass[a4paper,fleqn]{cas-sc}

\usepackage[numbers,sort]{natbib}
\usepackage{bm}
\usepackage{color}
\usepackage{subfigure}

\def\tsc#1{\csdef{#1}{\textsc{\lowercase{#1}}\xspace}}
\tsc{WGM}
\tsc{QE}

\begin{document}
\let\WriteBookmarks\relax
\def\floatpagepagefraction{1}
\def\textpagefraction{.001}

\shorttitle{Prompt2Gaussia}    

\shortauthors{S. Cui et al.}  

\title [mode = title]{Prompt2Gaussia: Uncertain Prompt-learning for  Script Event Prediction}  

\tnotemark[<tnote number] 


%
\author[a,b]{Shiyao Cui}
\ead{cuishiyao@iie.ac.cn}
\credit{Conceptualization, Methodology, Software, Writing - original draft}

\author[a,b]{Xin Cong}
\ead{congxin@iie.ac.cn}
\credit{Conceptualization, Formal analysis, Writing - review \& editing}

\author[a,b]{Jiawei Sheng}
\ead{shengjiawei@iie.ac.cn}
\credit{Validation,  Writing - review}

\author[a]{Xuebin Wang}
\ead{wangxuebin@iie.ac.cn}
\credit{Investigation}

\author[a,b]{Tingwen Liu}
\cormark[1]
\ead{liutingwen@iie.ac.cn}
\credit{Supervision, Funding acquisition, Project administration}

\author[c]{Jinqiao Shi}
\ead{shijinqiao@bupt.edu.cn}
\credit{Supervision}

\affiliation[a]{organization={Institute of Information Engineering, Chinese Academy of Sciences},
	city={Beijing},
	postcode={100190}, 
	country={China}}

\affiliation[b]{organization={School of Cyber Security, University of Chinese Academy of Sciences},
	city={Beijing},
	postcode={100049}, 
	country={China}}

\affiliation[c]{organization={School of Cyber Security, Beijing University of Posts and Telecommunications},
	city={Beijing},
	postcode={100088}, 
	country={China}}

\cortext[1]{Corresponding author}



\begin{abstract}
Script Event Prediction (SEP) aims to predict the subsequent event for a given event chain from a candidate list. Prior research has achieved great success by integrating external knowledge to enhance the semantics, but it is laborious to acquisite the appropriate knowledge resources and retrieve the script-related knowledge. In this paper, we regard public pre-trained language models as knowledge bases and  automatically mine the script-related knowledge via prompt-learning. Still, the scenario-diversity and label-ambiguity in scripts make it uncertain to construct the most functional prompt and label token in prompt learning, i.e., prompt-uncertainty and verbalizer-uncertainty. Considering the innate ability of Gaussian distribution to express uncertainty, we deploy the prompt tokens and label tokens as random variables following Gaussian distributions, where a prompt estimator and a verbalizer estimator are proposed to estimate their probabilistic representations instead of deterministic representations.  We take the lead to explore prompt-learning in SEP and provide a fresh perspective to enrich the script semantics.  Our method is evaluated on the most widely used benchmark and a newly proposed large-scale one. Experiments show that our method, which benefits from knowledge evoked from pre-trained language models, outperforms prior baselines by 1.46\% and 1.05\% on two benchmarks, respectively.
\end{abstract}




\maketitle

\section{Introduction}

\label{sec:intro}

\begin{figure}[h]
	\centering
	\includegraphics[width=0.95\linewidth]{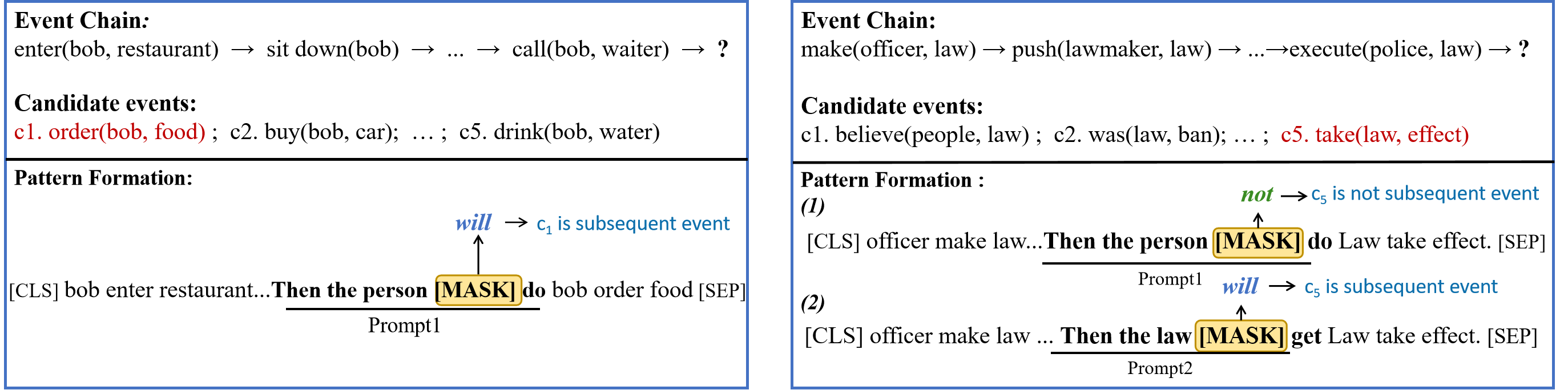}
	\caption{Two examples of SEP where the correct candidate event is marked in red. Due to the scenario-gap among scripts, the prompt working well for Example1 to infer the correct subsequent event fails  for Example2. }
	\label{fig:example}
\end{figure}

Script~\citep{Schank1977ScriptsPG} is a chain of events which describes the activities of a protagonist.
%
Given a sequence of ordered events, Script Event Prediction~(SEP) aims to predict the subsequent event from a candidate list.
%
As an important task to understand human behaviors and social development~\citep{EOM2021278,du-etal-2022-resin},   SEP involves scripts on various scenarios and has supported many applications including scenario dialogue systems~\citep{lv-etal-2020-integrating}, question answering~\citep{DBLP:conf/semweb/LiCWCHG19} and recommendation systems~\citep{CUI2023119007}.
Whereas, events in SEP are represented by a verb-centric tuple as Figure~\ref{fig:example} shows, where the absence of contexts makes the script semantics more sparse than normal texts, raising the difficulty of script understanding~\citep{Modi2016EventEF,DBLP:conf/sigir/ZhengCC20}.
Prior studies~\citep{lv-etal-2020-integrating,bai-etal-2021-integrating,Bai2022RichEM} have achieved great success by introducing external knowledge to enrich the semantics, but it is laborious to construct the appropriate knowledge bases and costly to retrieve the knowledge in need. 

Recent progress~\citep{10.1145/3560815} has revealed that pre-trained language models (PLMs) have acquired versatile factual knowledge from large-scale corpora. 
Particularly, the factual knowledge contains underlying principles which describe how events happen sequentially.
We therefore consider one question: \textit{can we probe the knowledge in PLMs, instead of retrieving it from knowledge bases, to improve SEP?}
The \textbf{prompt-learning}~\citep{DBLP:conf/nips/BrownMRSKDNSSAA20} points out a credible direction for us.
Prompt-learning seeks to elicit  knowledge from PLMs by transforming specific tasks into the same formation as the pre-training objective of PLMs.
Typically, given a masked language modeling (MLM) model $\mathcal{M}$ and its vocabulary $\mathcal{V}$, prompt-learning involves two key ingredients: 
(i) a pattern function, which designs a prompt to cast specific tasks as the MLM pattern; 
(ii) a verbalizer, which maps task labels into tokens in $\mathcal{V}$.
Prompt-learning achieves the specific tasks by predicting the \texttt{[MASK]} token in the prompt into the label token in the verbalizer.
By elaborately designing the functional \textbf{pattern-verbalizer pair} (\textbf{PVP})~\citep{schick-schutze-2021-exploiting}, prompt-learning bridges the gap between pre-training and fine-tuning, thus factual knowledge learned in the pre-training stage could be fully exploited to benefit downstream tasks~\citep{Ding2021PromptLearningFF,DBLP:conf/www/ChenZXDYTHSC22,ma-etal-2022-prompt,LI2022346}.

Motivated by this, we intend to adapt prompt-learning into SEP.
However, the PVP construction suffers from two kinds of \textbf{uncertainties} in SEP:
1) There exists \textit{scenario-diversity} in SEP, which means that different scripts may depict different real-world scenarios.
Since the prompt is highly sensitive to its surrounding contexts~\citep{shin-etal-2020-autoprompt}, the diverse scenarios need scenario-aware prompts to probe the scenario-related factual knowledge from PLMs.
For example in Figure~\ref{fig:example}, for the ``restaurant visiting'' scenario which depicts a series of behaviors of a person to have dinner, it needs a \textit{person}-involved prompt, ``\texttt{Then the person [MASK] do}'', to probe restaurant visiting knowledge.
However, such a \textit{person}-involved prompt fails for scripts on a ``law enforcement'' scenario to probe the \textit{law}-related knowledge.
Unfortunately, the tuple-based scripts only express uncertain and vague scenario semantics due to the absence of context in SEP.
It struggles to construct a certain prompt for each script based on the sparse scenario semantics in tuples.
We name this issue as \textbf{prompt-uncertainty}.
However, existing prompt construction methods~\citep{Liu2021GPTUT,li-liang-2021-prefix,liu-etal-2022-dynamic} deploy certain prompt tokens with deterministic representations, which could not be adapted into SEP directly to overcome such prompt-uncertainty.
2) A similar issue occurs in the verbalizer construction.
Normally, the verbalizer is built by the label-semantics-related words to drive the task predictions.
Unfortunately, in SEP, there exists \textit{label-ambiguity}. 
Specifically, the label ``\texttt{YES}'' could tag the correct candidate event,  but the semantics of this label is too ambiguous to decide the label-semantics-related words which could associate with the correct candidate event in prompt-learning.
As the associating relation between the event chain and the correct candidate is fine-grained and complicated~\citep{lee-goldwasser-2019-multi}, it is uncertain to select the best-performing label token to cover the complicated associating relation.
Although some prompt-learning studies propose to auto-search the label tokens~\citep{Schick2021ExploitingCF}, these methods still cast certain label tokens, thus struggling to model the semantic uncertainty.
Here, we name this issue as \textbf{verbalizer-uncertainty}.

In this paper, we propose a novel approach, Prompt2Gaussia (\textbf{P2G}),  which copes with the issues above via Gaussian estimation in prompt-learning. Specifically, instead of deterministic representations, a prompt estimator and a verbalizer estimator are designed to represent the prompt and label tokens as Gaussian distributional representations, where the \textit{mean}  and \textit{variance} of Gaussian embedding express the semantics and uncertainty, respectively. For the \textbf{prompt-uncertainty}, the prompt estimator first grasps the scenario clues of the script by attentively integrating the semantics of the event chain and candidate events. Then, to model the uncertainty of the ambiguous scenario semantics, the estimator approximates the prompt tokens as Gaussian embeddings. For the \textbf{verbalizer-uncertainty}, the verbalizer estimator exploits Gaussian embeddings of multiple learnable continuous tokens to conquer the uncertainty caused by the label-ambiguity. To integrate semantics from these label tokens, an uncertainty-aware semantics aggregation module is designed to weigh them for a more predictive final label token. Thanks to the Gaussian distributional representations of prompt and label tokens, P2G could overcome the problems of semantic uncertainty. In this way, P2G evokes knowledge from pre-trained language models and is thus free of external knowledge resources.
Overall, our contributions are as follows:
\begin{itemize}
	\item We propose a novel method, Prompt2Gaussia (P2G), to probe knowledge from PLMs for SEP. To our best knowledge, we take the lead to adapt prompt-learning in SEP. %
	\item We identify two types of uncertainties in SEP and deploy the prompt and label tokens as Gaussian embeddings to model such uncertainties.
	\item Experiments on two public benchmarks show that our method outperforms prior competitive baselines. Extensive experiments are conducted to analyze how our method works.
\end{itemize}

\section{Related Work}
\subsection{Script Event Prediction}
Script event prediction (SEP) is an important task to understand the principles of event evolution  and has drawn great research attention.
Existing studies could be grouped into two lines.
The \textbf{first} line of studies focus on modeling the event co-occurrence from four aspects. 
1) Early methods~\citep{chambers-jurafsky-2008-unsupervised, jans-etal-2012-skip, pichotta-mooney-2014-statistical,rudinger-etal-2015-script, GranrothWilding2016WhatHN,wang-etal-2017-integrating} model the semantic correlations between \textit{event-pairs} to infer the missing event.
2) Limited by the semantics of event-pair, researchers further explore the full \textit{event-chain} to conduct SEP.
Specifically, Lv et al.~\cite{Lv2019SAMNetIE} regard the given event chain as a combination of event-segments and exploit the segmental-level semantics for task prediction. 
For more fine-grained clues, the event chains are more explicitly modeled at the event-level ~\citep{Pichotta2016LearningSS,DBLP:journals/ijis/ZhouWWPYX22} or argument-level~\citep{Wang2021MultilevelCE}.
3) Since prior event-pair and event-chain based methods can not take advance of the dense event correlations, the \textit{event-graph}~\citep{Li2018ConstructingNE,zheng-etal-2020-heterogeneous,du-etal-2022-graph} based methods are proposed, where the graph structure is utilized to model the evolution principles of the script events.
4) Zhu et al.~\cite{Zhu2022AGA} predict
the next event with a \textit{generative paradigm} which is pre-trained with an event-centric objective to explore the event correlations.

Due to the sparse semantics of scripts, the \textbf{second} line of methods focus on integrating external knowledge for SEP.
Regarding of the introduced knowledge, these methods could be grouped into three groups.
1) The \textit{discourse relations} from Penn Discourse Tree Bank (PDTB)~\citep{Prasad2006ThePD} are explored by Lee et al. ~\cite{lee-goldwasser-2019-multi,DBLP:conf/emnlp/LeePG20} to guide the event prediction.
2) For more sufficient guidance, the \textit{commonsense knowledge bases}, which include Event2Mind~\citep{rashkin-etal-2018-event2mind}, ATOMIC~\citep{Sap2019ATOMICAA} and ASER~\citep{Zhang2020ASERAL}, are introduced by Ding et al.~\cite{ding-etal-2019-event-representation}, Zhou et al.~\cite{zhou-etal-2021-modeling} and Lv et al.~\cite{lv-etal-2020-integrating} to enrich the script semantics. 
Besides,  \textit{commonsense relations} from TimeTravel~\cite{qin-etal-2019-counterfactual} are introduced by Zhou et al.~\cite{10.1145/3459637.3482150}.
%
3) To precisely locate the script-related knowledge, the \textit{original texts} of the script events are traced by researchers~\citep{wang-etal-2021-incorporating-circumstances,bai-etal-2021-integrating,Bai2022RichEM}, where all the constituents in the texts serve to enhance the semantics.

In this paper, we regard the PLMs as extra knowledge bases via prompt-learning.
With the merits of prompt-learning, the script-related knowledge is probed simultaneously when SEP is formulated into a masked language modeling paradigm.
Hence, P2G is free of the external knowledge resources mentioned above, making this method more portable to conduct.
To the best of our knowledge, we are the first to probe knowledge from PLMs to boost the prediction to subsequent events in SEP.

\subsection{Prompt-learning}

GPT-3~\citep{DBLP:conf/nips/BrownMRSKDNSSAA20} inspires the emergence of prompt-learning, which exploits a pattern-verbalizer pair (PVP) functions to  cast downstream tasks into the same paradigms as the language model pre-training.
Specifically, the pattern function designs a prompt to re-pattern the downstream tasks, and the verbalizer maps the task labels as a set of specific tokens.
Early research exploits the PVP construction using the manually-designed prompt~\citep{Trinh2018ASM, davison-etal-2019-commonsense,schick-schutze-2021-just} and label tokens~\citep{yin-etal-2019-benchmarking, cui-etal-2021-template, schick-schutze-2021-exploiting}.
To avoid the consuming labor cost, researchers studied how to automatically search discrete tokens~\citep{gao-etal-2021-making} as the prompt and label tokens~\citep{Schick2020AutomaticallyIW, Schick2021ExploitingCF}.
Further, prompt~\citep{Liu2021GPTUT,Lester2021ThePO,liu-etal-2022-dynamic} and label tokens~\citep{hambardzumyan-etal-2021-warp,cui-etal-2022-prototypical} are deployed as learnable continuous embeddings to get rid of the designing of prompt and label tokens.
Despite these great efforts, existing methods treat prompt and label tokens as deterministic tokens in the semantic space, which suffer from the uncertainty to construct the best-performing PVP in SEP.

\subsection{Uncertainty Modeling in NLP}
In most Natural Language Processing (NLP) studies, the high-level representation of a token is modeled as a fixed-length feature vector, which could be regarded as a \textit{point} in the semantic space, namely the \textit{point embedding}~\citep{Qian2021ConceptualizedAC}.
However, the point embedding struggles to model the uncertainty among tokens, since it deploys one token as a deterministic point and assumes that the learned token features are always correct~\citep{Cao2021UncertainEFI}.
To alleviate such weaknesses, Vilnis et al.~\cite{Vilnis2015WordRV} propose Gaussian embedding to represent tokens under a soft region in the semantic space, where the covariance in Gaussia expresses the semantic uncertainty.
Modeling the semantic uncertainties using Gaussian embeddings has shown its sparkles in several NLP tasks~\cite{Xiao2019QuantifyingUI}, including named entity recognition and language modeling. To the best of our knowledge, we are the first to explore Gaussian embedding to model the uncertainty in prompt-learning for SEP.

\section{Preliminary}

\subsection{Problem Definition}

The goal of Script event prediction (SEP) task is to predict the subsequent event for a given event chain from a candidate list. 
Formally, each script instance contains an ordered $n$-event chain $\mathtt{E} = \left< e_1, e_2, ..., e_{n} \right>$ and an $m$-event candidate list $\{e_{c_1}, e_{c_2}, ..., e_{c_m}\}$, where SEP seeks to predict  $e_{n+1}$  to $\mathtt{E}$ from the candidates. 
In scripts, each event $e_i$ contains four arguments: subject $a_s$, verb $a_v$, object $a_o$ and indirect object $a_p$, and is denoted as a tuple $e_i = a_{v_i}(a_{s_i},a_{o_i},a_{p_i})$.
For example, an event tuple, $e_i = give(waiter,bob,food)$, means that ``waiter gives Bob food''. If $e_i$ lacks one argument, ``NULL'' is used to represent the empty argument.

\subsection{Prompt-learning}

Prompt-learning methods explore knowledge distributed in PLMs by transforming the specific tasks into the paradigm which holds the same training objective as language model pre-training.
For a pre-trained masked language model $\mathcal{M}$ and its vocabulary $\mathcal{V}$, prompt-learning employs $\mathcal{M}$ directly as a predictor through the completion of a cloze-style task.
Specifically, for a sequence $x$ and its label $y$, a pattern mapping function $f$ first adds a prompt including one masked token $\mathtt{[MASK]}$  to $x$, obtaining $f(x)$ in the pattern of prompt-learning.
Next, a verbalizer $v$ maps $y$ into a label token $v(y) \in \mathcal{V}$, and the MLM-head of $\mathcal{M}$ predicts the masked token  as the designated label token $v(y)$.
Accordingly, the probability to that  the input sequence $x$ is of $y$-type could be obtained as follows:
\begin{equation}
	\label{equ:prompt-example}
	P(y|x) = P\left( \mathtt{[MASK]} = v(y) | f(x) \right),
\end{equation}
where a classification problem is transformed into the masked language modeling (MLM) paradigm, which bridges the gap between $\mathcal{M}$'s pre-training and fine-tuning.

\begin{figure*}
	\centering
	\includegraphics[width=0.98\linewidth]{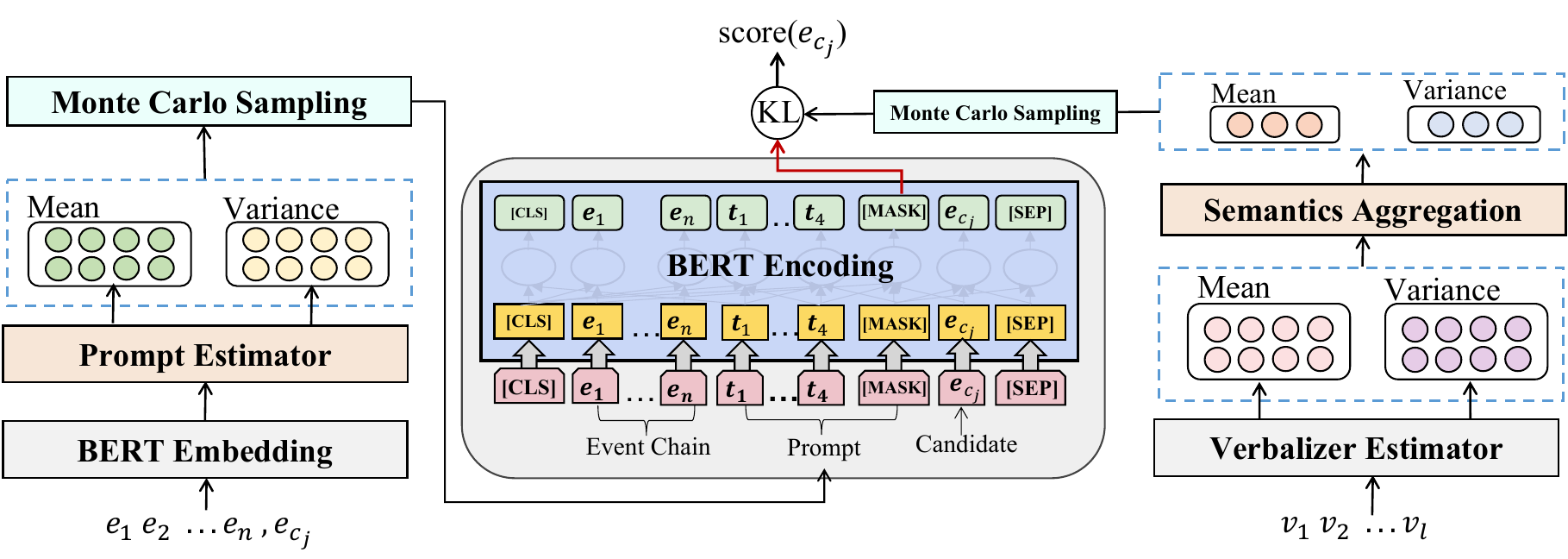}
	\caption{Model illustration, where $e_i$ / $e_{c_j}$ represents the specific event and $v_j$ denotes the learnable label token.  }
	\label{fig:model}
\end{figure*}

\section{Method}

Considering the great ability of masked language modeling (MLM) for context learning~\citep{shen-etal-2022-event}, we exploit MLM pre-trained language model in our method.
To elicit knowledge from PLMs, we need to design the well-performing pattern-verbalizer pair (PVP) for each script instance.
In this section, we detail our proposed PVP construction as Figure~\ref{fig:model} shows.

\subsection{Pattern Construction}

In pattern construction, a pattern function, $f$, reframes SEP into the MLM pattern.
Due to the  prompt-uncertainty, a prompt estimator is proposed to approximate the Gaussian distributional representations of prompt tokens.

\subsubsection{Pattern Formation}

We first introduce how the pattern function, $f$, formats SEP into the MLM pattern by adding a prompt for each script instance.
To query the language model $\mathcal{M}$ for clues to judge the correct subsequent event, $f$ respectively adds one prompt between each candidate event $\mathtt{e}_{c_j}$ and the event chain $\mathtt{E} = \left< \mathtt{e}_{1}, ..., \mathtt{e}_{n} \right>$. 
Accordingly, the pattern construction toward one candidate event and the event chain in a script instance could be specified as follows:
\begin{equation}
	\label{equ:prompt}
	\begin{aligned}
		f(e_{c_j}) = \mathtt{[CLS]~E}~\mathtt{[t]}_{j,1},...,  \mathtt{[t]}_{j,k} \mathtt{[MASK]} ~\mathtt{[e]}_{c_j}~\mathtt{[SEP]}, & \\
	\end{aligned}
\end{equation}
where $\mathtt{[t]}_{j,1},...,  \mathtt{[t]}_{j,k} \mathtt{[MASK]}$ refer to the added prompt.
The prompt contains $k+1$ tokens, where $\mathtt{[MASK]}$ works for the task prediction, and $\mathtt{[t]}_{j,1},..., \mathtt{[t]}_{j,k}$ are $k$ prompt tokens which are approximated as  Gaussian distributional embeddings in Section~\ref{sec:prompt_estimation} to model the prompt-uncertainty.

Note that in this process above, each event $\mathtt{e}_{i}$ / $\mathtt{e}_{c_j}$  is converted into the natural language texts ``$s_i \;  v_i \; o_i \; p_i $'' or ``$s_{c_j} \; v_{c_j} \; o_{c_j} \; p_{c_j}$''.
The pattern construction is respectively conducted towards each candidate event of the given event chain.
Then, we employ the embedding layer of $\mathcal{M}$ to convert the $\mathtt{[MASK]}$ token and arguments in each event into the real-valued representations for the following prompt estimation.

\subsubsection{Prompt Estimator}

\label{sec:prompt_estimation}

Due to the prompt-uncertainty, we treat each prompt token under a soft region following Gaussian distributions to grasp the semantic uncertainties~\citep{Vilnis2015WordRV, Qian2021ConceptualizedAC}.
To derive the prompt which is adaptive with each script scenario, we derive the Gaussian embeddings of the prompt tokens using the specific script events, so that the script-specific scenario semantics could be injected into the prompt. 

Specifically, considering that each argument in the  correct candidate event usually shows associations with the event chain~\citep{Wang2021MultilevelCE}, we derive the mean and variance of prompt Gaussian embeddings using the arguments (i.e., subject, verb, object, indirect object) in the candidate event.
Accordingly, four prompt tokens are employed in total (i.e., $k=4$ in Eq.~\ref{equ:prompt}), where each corresponds to one argument type.
Consequently, for each pattern $f(e_{c_j})$, we derive the prompt by refining argument-related information from the event chain using the argument in each candidate event $e_{c_j}$.
For example, we derive the mean $\bm{\mu}_{j,1} \in  \mathbb{R}^{1 \times d}$ and variance $\bm{\sigma}^2_{j,1} \in \mathbb{R}^{1 \times d}$ of the first prompt token  $  \mathtt{[t]}_{j,1} $  using its corresponding argument $a_{s_{c_j}}$ as follows:
\begin{equation}
	\label{equ:attn}
	\begin{aligned}
		& \bm{\mu}_{j,1}  = {\rm{Attn}}_{\mu}(\bm{a}_{s_{c_j}},  \bm{E},  \bm{E} )\\
		&  \bm{\sigma}^2_{j,1}  = \verb|exp| \left( {\rm{Attn}}_{\sigma}(\bm{a}_{s_{c_j}},  \bm{E},  \bm{E} ) \right),
		& 
	\end{aligned}
\end{equation}
where $\bm{a}_{s_{c_j}}\in\mathbb{R}^{1 \times d}$ is the subject embedding of event $e_{c_j}$, $\mathbf{E} \in \mathbb{R}^{ (4\times n) \times d} $ is the concatenation of $a_{s_i}~a_{v_i}~a_{o_i}~a_{p_i} $ embeddings of $n$ events in the event chain, and $d$ is the token embedding size.
``${\rm{Attn}}$'' refers to the scaled dot-product attention~\citep{NIPS2017_3f5ee243} with Query/Key/Value, where $\mathbf{a}_{s_{c_j}}$ works as Query, and $\mathbf{E}$ serves as the Key-Value pair.
Since the mean and variance describe the different aspects of the Gaussian distributional embeddings~\citep{DBLP:conf/aaai/ZhangZHY21}, a coupled $\verb|Attn|_{\mu}$ and $\verb|Attn|_{\sigma}$ serve to respectively derive the mean and variance vector.

For each candidate event $\bm{e}_{{c_j}}$, we leverage $\mathcal{N} ( \bm{\mu}_{j,1} ,  \bm{\sigma}^2_{j,1} )  $ to denote the Gaussian embedding of prompt token $  \mathtt{[t]}_{j,1} $.
For other prompt tokens   $  \mathtt{[t]}_{j,2}, \mathtt{[t]}_{j,3}, \mathtt{[t]}_{j,4} $ in $f(e_{c_j})$, we respectively derive their Gaussian distributional embeddings using $ a_{v_{c_j}}, a_{o_{c_j}}, a_{p_{c_j}} $ via the same procedure above.

\subsection{Verbalizer Construction}

The verbalizer builds the mapping between the task label $y$ into a token $v(y) \in \mathcal{V}$, so that the MLM-head of $\mathcal{M}$ could be used for specific task predictions.
Vanilla prompt-learning methods build the verbalizer based on expertise, but the ambiguity of the SEP label causes verbalizer-uncertainty, making it uncertain to construct the mapping $y \rightarrow v(y)$.
Therefore, instead of certain tokens with deterministic representations, we attempt to cast label tokens via Gaussian embedding in the continuous semantic space in $\mathcal{V}$ to model the semantic uncertainty.

\subsubsection{Verbalizer Estimator}

To learn the appropriate semantics for the label token $v(y)$, we first randomly initialize it as a few trainable parameters $ \bm{v} \in \mathbb{R}^{1 \times d}$.
Due to the semantic uncertainty from label-ambiguity, we represent the label token using Gaussian embedding.
Since prompt-learning aims to predict the label token from $\mathcal{M}$'s vocabulary $\mathcal{V}$, we derive the mean and variance of label token $v(y)$ using $\mathcal{M}$'s MLM-head, which associates the label token with the $\mathcal{V}$.
Specifically, the process above could be formulated as:
\begin{equation}
	\label{equ:verbalizer}
	\begin{aligned}
		& \bm{\mu}_v  = {\rm{MLM}\text{-}\rm{head}}\left( {\rm{ReLU}}( \bm{v} \bm{W}_{\mu} + \bm{b}_{\mu}) \right) \\
		& \bm{\sigma}^2_v  = {\rm{exp}} \left(  {\rm{MLM}\text{-}\rm{head}} \left( {\rm{ReLU}}( \bm{v} \bm{W}_{\sigma} + \bm{b}_{\sigma}) \right)  \right),
	\end{aligned}
\end{equation}
where $\bm{W}_{\mu}$, $\bm{W}_{\sigma}$, $\bm{b}_{\mu}$, $\bm{b}_{\sigma}$ are trainable parameters. 
$\mathcal{N} ( \bm{\mu}_v,  \bm{\sigma}_v^2 )$ represents the label token $v(y)$ with a soft region in the $\mathcal{V}$'s semantic space. 

\subsubsection{Uncertainty-aware Semantics Aggregation}
Due to the semantic limitation of one single token to express task labels~\citep{DBLP:journals/corr/abs-2210-12435}, we employ multiple tokens to construct the verbalizer for the complicated semantic relation between  candidate events and the event chain.
We therefore design this module to aggregate their semantics.

Supposing that a combination of $l$ tokens collaboratively serve as the label tokens, we first follow Eq.~\ref{equ:verbalizer} to acquire their Gaussian embeddings as $\mathcal{N}_1 ( \bm{\mu}_{v,1},  \bm{\sigma}_{v,1}^2 ) , ... ,  \mathcal{N}_l ( \bm{\mu}_{v,l},  \bm{\sigma}_{v,l}^2 )$.
As prior research~\citep{Feng2021UncertaintyawareAG,Cao2021UncertainEFI} suggested, a token with larger uncertainty may carry some unnecessary noises, we thus employ a smooth exponential function to balance the contribution of these label tokens on weight $\kappa_i$ as follows:
\begin{equation}
	\label{equ:para}
	\kappa_i = {\rm{exp}} ( - \lambda \bm{\sigma}_i ),
\end{equation}
where $\kappa_i$ is the weight for the $i_{\rm{th}}$ label token and $\lambda   $ is a hyper-parameter.
Consequently, we aggregate Gaussian embeddings of these tokens  as follows:
\begin{equation}
	\hat{ \bm{\mu} }_v  =\sum_{i=1}^{l} (\kappa_i \odot \bm{\mu}_i )   \quad \hat{ \bm{\sigma} }_v^2  =\sum_{i=1}^{l} (\kappa_i \odot \kappa_i \odot \bm{\sigma}_i^2 ) ,
\end{equation}
where $\odot$ denotes the element-wise product. 
Finally, $ \mathcal{N}( \hat{ \bm{\mu}_v },  \hat{\bm{\sigma}}_v^2 ) $ is Gaussian embedding of the final label token $v(y)$. 

\subsection{Prediction}

To predict the subsequent event, we compute each candidate event's score $\verb|score|(e_{c_j})$ to justify whether it is the correct subsequent event.
According to Eq.~\ref{equ:prompt-example}, $\verb|score|(e_{c_j})$ equals to the probability of how $v(y)$ could fill the masked token $\mathtt{[MASK]}$ in the pattern $f(e_{c_j})$ as follows: 
\begin{equation}
	\label{equ:score}
	{\rm{score}}(e_{c_j}) =  P\left(\mathtt{[MASK]}=v(y)| f(e_{c_j})\right).
\end{equation}

Since all prompt tokens are represented as Gaussian distributional embeddings, we obtain their explicit representations using  Monte Carlo Sampling~\citep{gordon2018metalearning} technique. 
To make the sampling process differentiable, we further adopt the  re-parameterization trick~\citep{Kingma2014AutoEncodingVB}.
We illustrate this process by taking  the first prompt token $\mathtt{[t]}_{j,1}$, whose Gaussian embedding is $\mathcal{N}(\bm{\mu}_{j, 1}, \bm{\sigma}^2_{j, 1})$ for $j_{th}$ candidate event, as an example.
Specifically, its representation could be obtained as follows:
\begin{equation}
	\bm{z}_{j, 1} =  \bm{\mu}_{j, 1} + \epsilon \bm{\sigma}_{j, 1}, \text{where} ~ \epsilon  \sim \mathcal{N} (0, 1),
\end{equation}
where $(\bm{\mu}_{j, 1}, \bm{\sigma}^2_{j, 1})$ are acquired from Eq.~\ref{equ:attn} and $\epsilon$ is a random noise from the standard Gaussian distribution.
The representations of other prompt tokens are derived in the same process above.
After obtaining the representations of prompt tokens, we feed $f(e_{c_j})$ into $\mathcal{M}$, where the $\rm{MLM}\text{-}\rm{head}$ of $\mathcal{M}$ could produce the logits, $\mathcal{M}(f(e_{c_j}))$, towards the vocabulary.
Next, a softmax layer over  $\mathcal{M}(f(e_{c_j}))$  yields the probability distribution $\bm{p}_j \in \mathbb{R}^{1 \times |\mathcal{V}| }$ over the vocabulary.
The process could be formulated  as follows:
\begin{equation}
	\label{equ:mlm-prob}
	\bm{p}_j =  {\rm{softmax}} \left( \mathcal{M}\left( f(e_{c_j}) \right) \right),
\end{equation}
where $\bm{p}_j$ indicates how each token in $\mathcal{V}$ is likely to fill $\mathtt{[MASK]}$.

Then, we derive the label token, whose Gaussian embedding is $ \mathcal{N}( \hat{ \bm{\mu}_v },  \hat{\bm{\sigma}}_v^2 ) $, from the verbalizer.
Specifically, we first sample the explicit label token representation $ \bm{v}_v \in \mathbb{R}^{1 \times |\mathcal{V}|}$ from its Gaussian embedding by employing the same sampling process as follows:
\begin{equation}
	\bm{v}_v  =  \hat{ \bm{\mu}_v} + \epsilon \hat{ \bm{\sigma}_v}, \text{where} ~ \epsilon  \sim \mathcal{N} (0, 1).
\end{equation}
We then normalize $\bm{v}_v$ with the softmax function to obtain the probability $\bm{p}_v \in \mathbb{R}^{1 \times |\mathcal{V}|}$ over the vocabulary:
\begin{equation}
	\label{equ:vocab-prob}
	\bm{p}_v = {\rm{softmax}}(\bm{v}_v)
\end{equation}
where each element of $\bm{p}_v$ denotes the semantic similarity between the label token and the corresponding token in $\mathcal{V}$.

Since that $\bm{p}_j$ and $\bm{p}_v$ are both distributions over $\mathcal{V}$, we employ Kullback–Leibler (KL) divergence to measure the semantic relevance between the label token and the predicted token of $\mathtt{[MASK]}$. The relevance score expresses how the label token $v(y)$ is likely to fill the masked position.
Correspondingly, Eq.~\ref{equ:score} could be further computed as follows:
\begin{equation}
	\label{equ:prob}
	\begin{aligned}
		{\rm{score}}(e_{c_j}) & =  P(\mathtt{[MASK]}=v(y)| f(e_{c_j})) \\
		& =  \frac{\texttt{exp}\left( \mathbb{D}_{KL}(\bm{p}_j~||~\bm{p}_v) \right)}{\sum_{k=1}^{m}{  \texttt{exp}\left( \mathbb{D}_{KL}(\bm{p}_k~||~\bm{p}_v ) \right)}},
	\end{aligned}
\end{equation}
where ${\rm{score}}(e_{c_j})$ denotes how the $j_{th}$ candidate event $e_{c_j}$ could be the correct candidate event.
The candidate event, which receives the highest score, is chosen as the final subsequent of the given event chain.

\subsection{Model Training}
Given all training instances where each contains an event chain and  $m$ candidates, we minimize the negative log-likelihood loss of the ground truth candidate event as follows:
\begin{equation}
	\mathcal{L}(\Theta) = - \sum^{N}_{t=1}{\verb| log|({\rm{Score}}_{t,j})}
\end{equation}
where $N$ is the number of the training instances, ${\rm{Score}}_{t,j}$ equals  ${\rm{score}}(e_{c_j})$ for the $j_{\rm{th}}$ candidate event of the $t_{\rm{th}}$ training instance, and the $j_{\rm{th}}$ candidate event is the ground truth. $\Theta$ is the set of all trainable model parameters.

\section{Experiment}

\subsection{Dataset and Evaluation}

We conduct experiments with \textbf{two} public benchmarks.
The \textbf{first} benchmark, which was  made public by Li et al.~\cite{Li2018ConstructingNE}, is the most widely used benchmark for SEP evaluation~\cite{Lv2019SAMNetIE,Li2018ConstructingNE,Wang2021MultilevelCE,du-etal-2022-graph}.
The event chains are extracted from  the New York Times (NYT) portion of the Gigaword corpus~\cite{Graff2003}.
Each event tuple is extracted by the C\&C tools~\cite{curran-etal-2007-linguistically} used for POS tagging and dependency parsing, and OpenNLP works for phrase structure parsing and coreference resolution.
%
%
The \textbf{second} benchmark was proposed and made public by Bai et al.~\cite{bai-etal-2021-integrating}  and Zhu et al.~\cite{Zhu2022AGA}.
With the event chain extraction code released by Granroth et al.~\cite{GranrothWilding2016WhatHN}, this benchmark was constructed upon the original document of the New York Times (NYT) portion of the Gigaword corpus by reproducing the event chain extraction process including  pos tagging, dependency parsing, and coreference resolution.
%
%
%
%
%
Considering the scale of the two benchmarks, we denote them respectively as ``\textbf{base benchmark}'' and ``\textbf{large benchmark}''.
Following prior research~\citep{Li2018ConstructingNE,bai-etal-2021-integrating}, we divide the training/development/test set and show the details in Table~\ref{tab:dataset}.
In both benchmarks, each script instance contains an event chain with 8 events and 5 candidate events, where only one candidate is correct.

For evaluation, we use Accuracy(\%), which is also adopted by our prior researchers~\cite{Li2018ConstructingNE,bai-etal-2021-integrating}, as the metric.

\begin{table}[t]
	
	\caption{The data statistics of the used base benchmark and large benchmark. }
	\setlength\tabcolsep{15pt}{
		\begin{tabular}{cccc}
			\toprule
			& Train set & Dev set & Test set \\
			\toprule
			Base benchmark & 140,331 & 10,000 & 10,000 \\
			Large Benchmark & 1,440,295 &  10,000  & 10,000 \\
			\bottomrule
	\end{tabular}}
	\label{tab:dataset}
\end{table}

\subsection{Implementation Details} 
We employ BERT-base-uncased~\citep{devlin-etal-2019-bert} as the pre-trained language model $\mathcal{M}$, and manually tune all the hyperparameters on the development set. 
Specifically, the hidden size of the all variables and representations are 768.
We utilize AdamW~\citep{Loshchilov2019DecoupledWD} optimizer with the learning rate of $1e-5$ to fine-tune $\mathcal{M}$ and the learning rate of $3e-5$ to optimize other  parameters.
We adopt L2 regularization to avoid overfitting and set its coefficient to $1e-8$.
For the base benchmark, we set batch size as $32$ and train our model for 10, 000 steps.
For the large barchmark, we set batch size as $32$ and train our model for 100, 000 steps.
The number of prompt tokens is $4$, and the number of label tokens is $3$.
We experiment using PyTorch 1.0.1  on the Nvidia Tesla V100 GPU. 
The implementation \textbf{source code} is available for the reproductivity check: \url{https://www.dropbox.com/s/l8cncfpaifdo0th/SEPcode.zip?dl=0}.

\subsection{Baselines}
We compare P2G against several typical SEP methods which could be categorized into three groups.

\textbf{Methods without extra knowledge} model the event co-occurrence to infer the subsequent event.
For the base benchmark, the baselines include
1) \textbf{EventComp}~\citep{GranrothWilding2016WhatHN} measures the event-pair similarity for prediction.
2) \textbf{PairLSTM}~\citep{wang-etal-2017-integrating} models the pairwise event relations for prediction.
3) \textbf{SGNN}~\citep{Li2018ConstructingNE}  explores the rich connections between script events using a graph structure. 
4) \textbf{SAM-Net}~\citep{Lv2019SAMNetIE} explores self-attention  upon diverse event-segments to infer the event.
5) \textbf{MCer}~\citep{Wang2021MultilevelCE} models the fine-grained connections between events within the event chain.
6) \textbf{GraphBERT}~\citep{du-etal-2022-graph} incorporates the structural event correlations of an event graph into BERT for event prediction.
For the newly proposed large benchmark, the baselines additionally include 
1) \textbf{MCPredictor-s}~\citep{bai-etal-2021-integrating} integrates the multiple narrative event chains of protagonists involved in the script for  prediction.
2) \textbf{SCPredictor-s}~\citep{bai-etal-2021-integrating}, a variant of MCPredictor-s, only explores the given narrative event chain for prediction.

\textbf{Methods with extra knowledge} introduce external knowledge to provide clues for the subsequent event prediction. 
For the base benchmark, the baselines include 1) \textbf{FEEL}~\citep{DBLP:conf/aaai/LeeG18} explores the sentiments of events and
the animacies of event participants to guide the task prediction.
2) \textbf{SGNN+Int+Senti}~\citep{ding-etal-2019-event-representation} leverages the knowledge about event intent and sentiment to enhance SGNN~\citep{Li2018ConstructingNE}; 
3) \textbf{ASER-Enhancement}~\citep{lv-etal-2020-integrating} incorporates eventuality knowledge graph ASER (activities, states, events, and their relations)~\citep{Zhang2020ASERAL} to predict the subsequent event.
4) \textbf{KG-Model}~\citep{zhou-etal-2021-modeling} explores the supportive event triples in the knowledge graph with similar inferential patterns.
For the newly proposed large benchmark, the baselines include
1) \textbf{MCPredictor}~\citep{bai-etal-2021-integrating} incorporates the original sentences of each event to enhance \textbf{MCPredictor-s}.
2) \textbf{SCpredictor}~\citep{bai-etal-2021-integrating} explores the original sentences of each event to enhance \textbf{SCPredictor-s}. 
3) \textbf{REP}~\citep{Bai2022RichEM} explores the rich event description parsed from  Abstract Meaning Representation (AMR)~\citep{banarescu-etal-2013-abstract} to boost the event prediction.

\begin{table}[t]
	\centering
	\caption{
		Overall results on the base benchmark, where the Wilcoxons test shows the significant difference (p < 0.05) between P2G and the strongest baseline Learnable Prompt + LR (then).}
	\setlength\tabcolsep{16pt}{
		\begin{tabular}{clc}
			\toprule
			\textbf{Category} & \textbf{Method} & \textbf{Accuracy(\%)}\\
			\midrule
			\multirow{6}{*}{w/o external knowledge}  & EventComp~\citep{GranrothWilding2016WhatHN} & 49.57 \\
			& PairLSTM~\citep{wang-etal-2017-integrating} & 50.83 \\
			& SGNN~\citep{Li2018ConstructingNE} & 52.45 \\
			& SAM-Net~\citep{Lv2019SAMNetIE} & 54.48 \\
			& MCer~\citep{Wang2021MultilevelCE} & 56.64 \\
			& GraphBERT~\citep{du-etal-2022-graph} & 60.72 \\
			\midrule
			\multirow{5}{*}{w/ external knowledge} & FEEL~\citep{DBLP:conf/aaai/LeeG18} & 55.03 \\
			& SGNN+Int+Senti~\citep{ding-etal-2019-event-representation} & 56.03 \\
			& ASER Enhancement~\citep{lv-etal-2020-integrating} & 58.66 \\
			& KGModel~\citep{zhou-etal-2021-modeling} & 59.99 \\
			\midrule
			\multirow{4}{*}{Prompt-learning} & Manual PVP & 60.24  \\
			& Learnable Prompt + LR(then) & 60.32 \\
			& Learnable Prompt + LR(preferred) & 59.99\\
			& Learnable Prompt + LR(added) & 59.95 \\
			\midrule
			Ours & \textbf{P2G} & \textbf{61.78} \\
			\bottomrule
		\end{tabular}
	}
	\label{tab:small-performance}
\end{table}

\begin{table}[h]
	\centering
	\caption{
		Overall results on the large benchmark, where the Wilcoxons test shows the significant difference (p < 0.05)  between P2G and the strongest baseline Learnable Prompt + LR (preferred)}
	\setlength\tabcolsep{16pt}{
		\begin{tabular}{clc}
			\toprule
			\textbf{Category} & \textbf{Method} & \textbf{Accuracy(\%)}\\
			\midrule
			\multirow{6}{*}{w/o external knowledge}  			& EventComp~\citep{GranrothWilding2016WhatHN} & 50.19 \\
			& PairLSTM~\citep{wang-etal-2017-integrating} & 50.32 \\
			& SGNN~\citep{Li2018ConstructingNE} & 52.30 \\
			& SAM-Net~\citep{Lv2019SAMNetIE} & 55.60 \\
			& SCPredictor-s~\citep{bai-etal-2021-integrating} & 58.79 \\
			& MCPredictor-s~\citep{bai-etal-2021-integrating} & 59.24 \\
			\midrule
			\multirow{3}{*}{w/ external knowledge} 
			& SCPredictor~\citep{bai-etal-2021-integrating} & 66.24 \\
			& MCPredictor~\citep{bai-etal-2021-integrating} & 67.05\\
			& REP~\citep{Bai2022RichEM} & 60.08 \\
			\midrule
			\multirow{4}{*}{Prompt-learning} & Manual PVP & 66.90  \\
			& Learnable Prompt + LR(then) & 66.78 \\
			& Learnable Prompt + LR(preferred) & 67.11 \\
			& Learnable Prompt + LR(added) &  66.66\\
			\midrule
			Ours & \textbf{P2G} & \textbf{68.16} \\
			\bottomrule
		\end{tabular}
	}
	\label{tab:large-performance}
\end{table}

\textbf{Prompt-learning-based methods.} We employ manually designed and auto-constructed pattern-verbalizer pair (PVP) as prompt-learning-based baselines.
1) \textbf{Manual PVP}: We handcraft a series of PVP-pair using specific tokens and leverage the best-performing one as the baseline.
Specifically, the best PVP deploys ``\texttt{the next event is [MASK]}'' as prompt and uses the token ``\texttt{then}'' to indicate the correct candidate event.
2) \textbf{Learnable Prompt + LR}: We deploy prompt tokens as learnable parameters~\citep{Liu2021GPTUT}, and simultaneously, adopt the discrete token selection criterion, Likelihood Ratio (LR)~\citep{Schick2020AutomaticallyIW,gao-etal-2021-making}, to  auto-search specific label tokens.
The parameterized prompt are optimized during model training, and LR selects ``\texttt{then}'', ``\texttt{preferred}'' and ``\texttt{added}'' as the top 3 most appropriate label tokens.
We report the results with these three label tokens and the learnable prompt.

Note that 1) the results of EventComp, PairLSTM, SGNN and SAM-Net upon the large benchmark are reproduced by Bai et al.~\cite{bai-etal-2021-integrating}; 2) the prompt-learning baselines are realized by ourselves; 3) results of other baselines are directly adopted from the original papers for a fair comparison.

\subsection{Main Results}

The performances of P2G and baselines are shown in Table~\ref{tab:small-performance} and Table~\ref{tab:large-performance}, and P2G outperforms all competitive baselines. 
Though prior researchers~\citep{ding-etal-2019-event-representation,DBLP:conf/aaai/ZhangZHY21} mentioned that
1\% of accuracy improvement in SEP is challenging, P2G outperforms
the strongest baseline by 1.46\% and 1.05\% on two benchmarks, respectively. This phenomenon reveals
the excellence of our proposed P2G method.
Specifically, we have detailed observations and analysis as follows.

1) \textbf{Integrating extra knowledge gives a significant boost for SEP performances. } Though conventional methods design complicated models, most of them generally lag behind the methods enhanced with extra knowledge. 
Typically, SGNN+Int+Senti excels SGNN definitely by 3.58\% in Table~\ref{tab:small-performance}.
This reflects that extra knowledge, which provides rich semantic clues to the script, could substantially help the event prediction and thus benefit the SEP performance.

2) \textbf{Prompt-learning is powerful to elicit and retrieve inherent knowledge distributed in PLMs.} Despite of the promising performances of  methods with external knowledge, the prompt-learning-based methods achieve competitive performances without external resources.  This  answers the question we proposed in Introduction that prompt-learning could effectively explore knowledge distributed in PLMs to boost model performances of SEP.

3) \textbf{Modeling the uncertainty of PVP construction in SEP brings stable performance improvements over other prompt-learning baselines.} 
P2G simultaneously outperforms prompt-learning methods with the best-performing manually designed PVP and auto-searched PVP.
We owe this remarkable performance gain to that our method takes advantage of modeling the prompt-uncertainty and verbalizer-uncertainty.
Detailed analyses are conducted in Section~\ref{sec:analysis}.

\section{Analysis and Discussion}

\label{sec:analysis}

\subsection{Ablation Study}

\begin{table*}[h]
	\centering
	\caption{
		Ablation study, where PE and VE are short for prompt estimator and verbalizer estimator.
	}
	\setlength\tabcolsep{15pt}{
		\begin{tabular}{lcccc}
			\toprule
			\multirow{2}{*}{\textbf{Method}} & \multicolumn{2}{c}{\textbf{Base Benchmark}} & \multicolumn{2}{c}{\textbf{Large Benchmark}} \\
			& Accuracy(\%) & $\Delta$(\%) & Accuracy(\%) & $\Delta$(\%) \\
			\toprule
			\textbf{P2G} & \bf 61.78 & - & \bf 68.16 & -  \\
			\quad -- Uncertain Modeling in the PE &  60.82 & -0.96 & 67.37 & -0.79  \\
			\quad -- Uncertain Modeling in the VE &  61.12 & -0.66 & 67.29 & -0.87 \\
			\quad -- Uncertain Modeling in both PE \& VE &  60.72 & -1.06 & 67.08 & -1.08 \\
			\quad -- Scenario-aware prompt &  61.12 & -0.66 & 67.05 & -1.11 \\
			\quad -- Uncertainty-aware Semantic Aggregation &  61.23 & -0.55 & 67.16 & -1.00 \\
			\quad -- Multi label tokens & 61.08 & -0.70 & 67.56 & -0.80 \\
			\bottomrule
	\end{tabular}}
	\label{tab:ablation}
\end{table*}

\label{sec:ablation}
To investigate how each component in P2G contributes, we conduct an ablation study and present the results in Table~\ref{tab:ablation}.
Detailed observations and analysis are as follows.

1) --~Uncertainty modeling in the prompt estimator: To study whether modeling the uncertainty of the prompt helps to boost performances, we remove the variance vector in Eq.~\ref{equ:attn}  and  directly utilize the mean vector as deterministic prompt representations.
The result degradation reflects the importance of prompt uncertainty modeling in SEP. 

2) --~Uncertainty modeling in the verbalizer estimator: To certify the effectiveness of modeling verbalizer-uncertainty, we remove the variance vector in Eq.~\ref{equ:verbalizer} but directly leverage the mean vector, which expresses a deterministic point in the semantic space, as the label token representation.
The drop in results reveals that modeling the verbalizer uncertainty is indispensable.

3) --~Uncertainty modeling in prompt estimator (PE) and verbalizer estimator (VE): When we simultaneously remove the uncertainty modeling in PE and VE, the performances hurt.
This obvious degradation demonstrates that modeling the uncertainty indeed empowers the prompt-learning in SEP.

4) --~Scenario-aware prompt: To probe the necessity of exploiting scenario semantics in the prompt estimator, we deploy the mean and variance in PE as learnable parameters which are universal for all script instances.
In this situation, no scenario-aware semantics are injected into the prompt tokens.
The declination of performance signifies the importance of scenario information and reveals that a prompt equipped with scenario semantics could fit SEP better.

5) --~Uncertainty-aware Semantic Aggregation: When we remove the uncertainty-aware semantics aggregation module for multiple label tokens but directly sum their Gaussian embeddings, the performances degrade.
This phenomenon illustrates that it is necessary to fuse the label tokens with filtering their hidden noisy semantics.

6) --~Multiple label tokens: When we build the verbalizer using only one label token, the results decline. 
This signifies that multiple label tokens could model the complicated semantics between the event chain and subsequent events better.
Despite the relatively weak performance with one label token, it also surpasses prompt-learning-based baselines which also adopt only one label token.
This phenomenon again validates the superiority of our method.

\begin{table}[t]
	\caption{Analysis to the sampling times ($n$) where performances with different values of $n$ are deployed and Acc is short for Accuracy. }
	\setlength\tabcolsep{17pt}{
		\begin{tabular}{cccccc}
			\toprule
			\multirow{2}{*}{n} & \multicolumn{2}{c}{\textbf{Base Benchmark}} & \multicolumn{2}{c}{\textbf{Large Benchmark}} \\
			& Accuracy(\%) & $\Delta$(\%) &  Accuracy(\%) & $\Delta$(\%)  \\
			\toprule
			0 & 60.72 & --  & 67.28 & --   \\
			1  & 61.78 & +1.06 & 68.16 & +0.88 \\
			2  & 61.89 & +1.17 & 68.33 & +1.05 \\
			4  & 62.31 & +1.59 & 68.61 &  +1.33    \\
			6  &  62.64 & +1.92 & 68.89 & +1.61 \\
			8   &  62.83 & +2.11 & 69.11 & +1.83 \\
			10  & 63.10 & +2.38 &  69.23 & +1.95 \\
			\bottomrule
	\end{tabular}}
	\label{tab:sampling}
\end{table}

\subsection{Analysis to the Uncertain Modeling}
In our method, for the uncertainty modeling, Monte Carlo sampling approximates the marginalization to the Gaussian embeddings and derives the explicit prompt and label token representations.
Here, the increasing sampling times could probe more potential semantic points and thus the semantic uncertainty could be modeled more sufficiently. 
To verify the effectiveness of the uncertainty modeling, we conduct analysis by exploring the sampling times.

Specifically, we sample $n \in \{0, 1, 2, 4, 6, 8, 10\}$ respectively.
When $n = 0$, the mean of Gaussian distributional embeddings are directly adopted as the token representations without considering the variance.
In this situation, the semantic uncertainty is not modeled.
When $n > 0$, for each candidate event, each prompt and label token are sampled for $n$  representations.
Here, the $n$ sampled prompt representations are fed into $\mathcal{M}$ and $n$ logits $\mathcal{M}(f(e_{c_j}))$  are produced in Eq.~\ref{equ:mlm-prob}.
These $n$ logits are mean-pooled for the following softmax layer to derive the probability $\bm{p}_j$.
Similarly, the $n$ sampled label token representations are mean-pooled to derive $\bm{p}_v$ in Eq.~\ref{equ:vocab-prob} for predictions.
Correspondingly, we report the model performances upon the test set in Table~\ref{tab:sampling}.
Reading from the table, the increasing sampling times could obviously strengthen the model performances compared with $n=0$.
This demonstrates the effectiveness of uncertainty modeling, where more appropriate and comprehensive semantics are captured to overcome the uncertainty issues.

Note that the increasing sampling times would cost more inference time.
Specifically,  we notice that the time-consumption linearly increases with the sampling times $n (n \geq 1)$, since the computation in Eq.~\ref{equ:mlm-prob} and Eq.~\ref{equ:vocab-prob} are correspondingly conducted $n$ times.
Despite the increment of time cost, our model enjoys good flexibility, where we could decide the sampling times according to the trade-off between efficiency and performance.
Notably, the model performances of $n=1$ are excellent enough to surpass prior baselines.
Fortunately, the increment of time-consumption from $n = 0$ to $n = 1$ is moderate since sampling only once will not bring other extra computation except the re-parameterization in Eq.~\ref{equ:mlm-prob} and Eq.~\ref{equ:vocab-prob}.
We think that $n=1$ is a good balance point between efficiency and performance and thus we conduct the main experiment and all analyses with $n=1$.

\subsection{Case Study}

\begin{figure*}[t]
	\centering
	\includegraphics[width=0.98\linewidth]{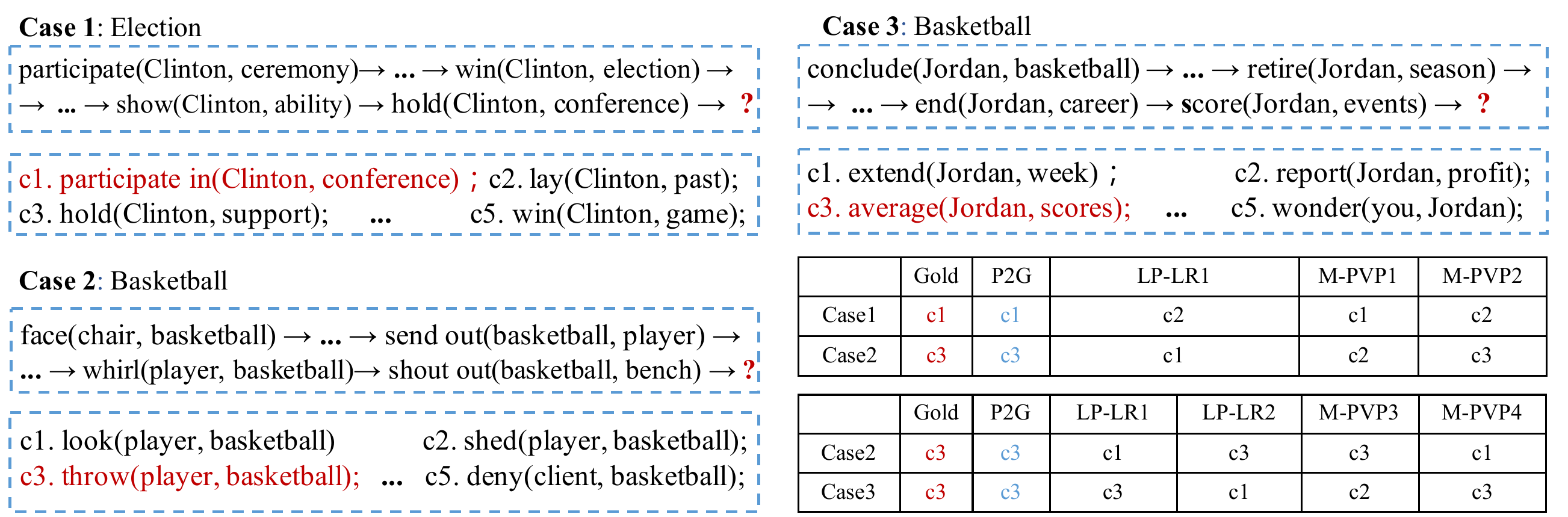}
	\caption{Case Study}
	\label{fig:case}
\end{figure*}

To illustrate how P2G benefits from the uncertainty modeling, we employ two pairs of cases in Figure~\ref{fig:case} for a case study.
Due to the space limitation, we only show associated events and candidates.

Specifically, Case1 and Case2 respectively show two scripts on ``election'' and ``basketball'' scenarios.
From them, we can see that the tuple-based event chains only provide uncertain and ambiguous scenario information.
For example, with the absence of the contexts, it is hard to understand why Clinton shows his ability after he won the election and why the basketball shout out the bench.
Based on the vague event chains, it is uncertain to construct the precise scenario-aware prompt.
Of course, we could customize prompts for these two cases based on human experitise. 
For hand-crafted PVP, the Manual-PVP1 (M-PVP1), whose prompt is ``\texttt{then the person do [MASK]}'', correctly infers the subsequent event for Case1 but fails for Case2.
Since the script in the election scenario focuses on the behaviors of a specific ``person'', the  prompt emphasizing ``person'' works well for it.
Meanwhile, the script in Case2 is not directly associated with an explicit person but the correlation between events, thus Manual-PVP2 (M-PVP2), whose prompt is ``\texttt{the next event is [MASK]}'', only works for Case2.
These phenomena manifest that scripts require scenario-aware prompt to make correct predictions but it is impossible to customize prompts for every script.
``Learnable Prompt + LR(then)'' (LP-LR1) learns the prompt automatically but fails to simultaneously choose the correct subsequent event for both cases.
It is because LP-LR1 only learned certain prompt which struggles to capture the scenario semantics from the uncertain event chains.
Thanks to the prompt-uncertainty modeling, P2G successfully works for two the scripts.

\begin{figure}[t]
	\centering
	\subfigure[Performance variances on the base benchmark]{
		\begin{minipage}[t]{0.48\textwidth}
			\centering
			\includegraphics[width=6cm]{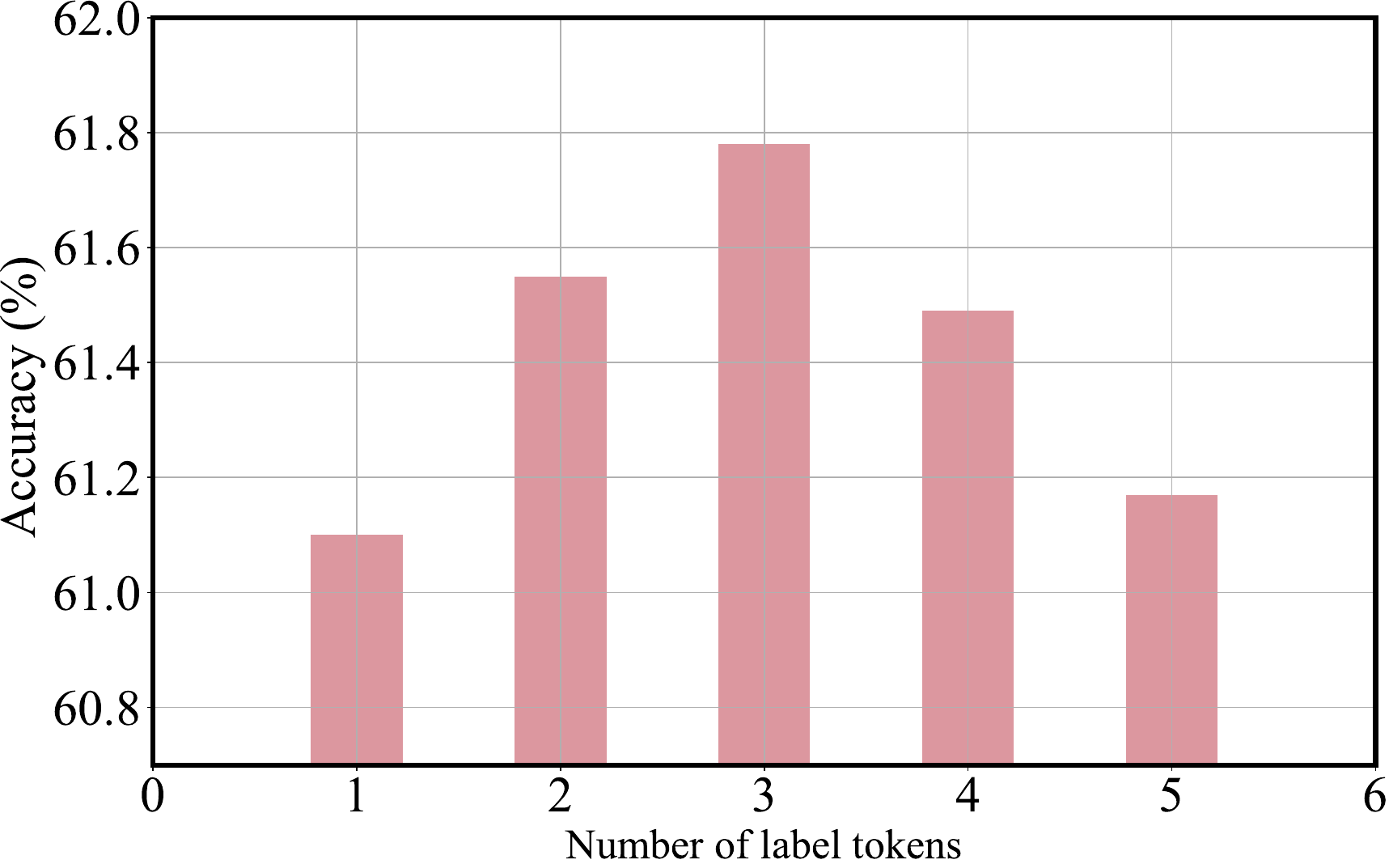}
	\end{minipage}}
	\subfigure[Performance variances on the large benchmark]{
	\begin{minipage}[t]{0.48\textwidth}
		\centering
		\includegraphics[width=6cm]{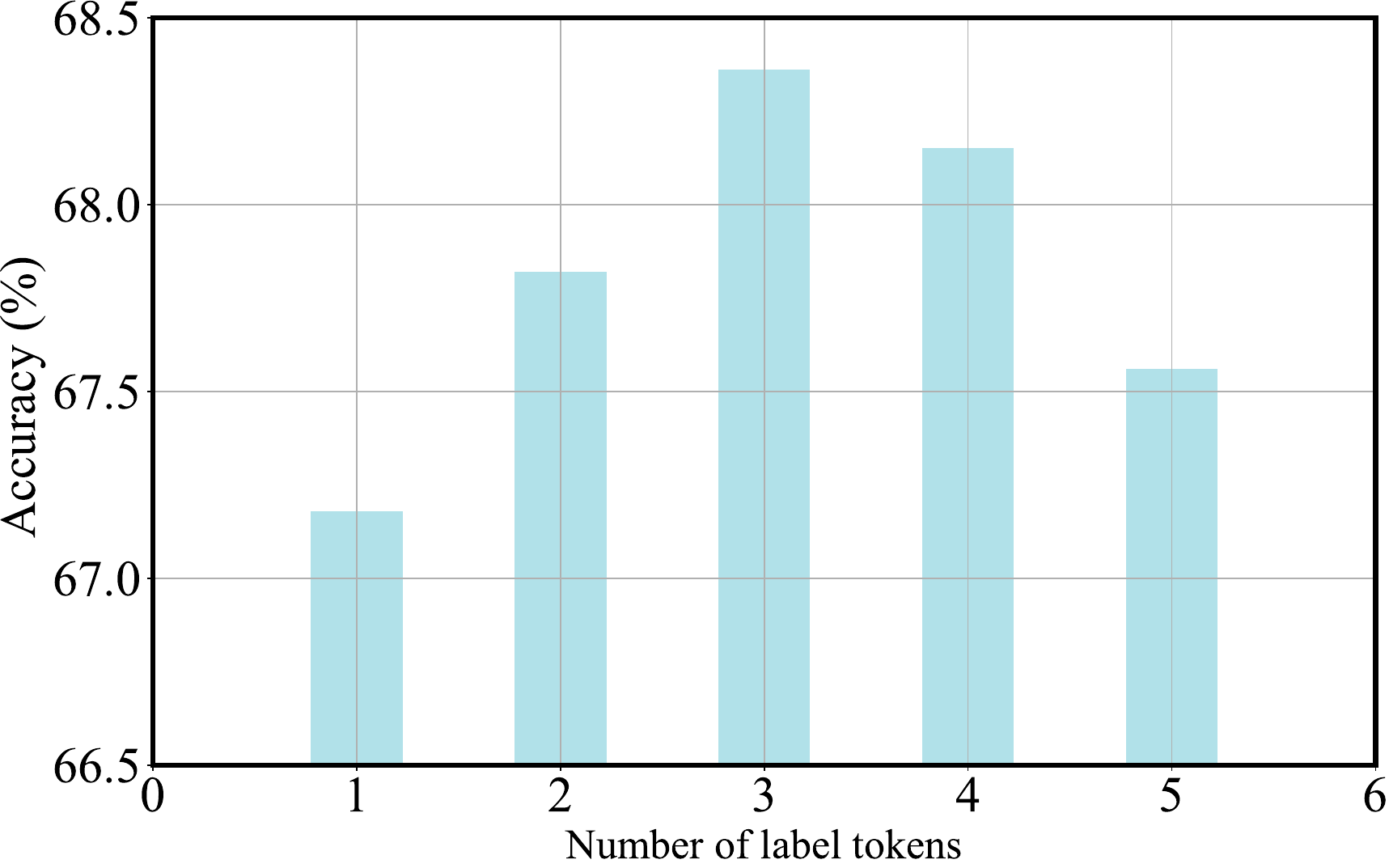}
	\end{minipage}}
	\caption{Performance variances with the number of label tokens on the base and large benchmark.}
	\label{fig:tokens}
\end{figure}

Case2 and Case3 illustrate two scripts which suffer from verbalizer uncertainty.
Specifically, for each script, Manual-PVP3 and Manual-PVP4  respectively show different performances with different label tokens (``\texttt{then}'' / ``\texttt{added}''), though their prompt keeps the same (``\texttt{the next event is [MASK]}'').
Similarly, performances of ``Learnable Prompt + LR(then)''  and ``Learnable Prompt + LR(added)'' also vary with their label tokens.
These phenomena reveal the uncertainty to choose the functional label token for different scripts.
Meanwhile, P2G  employs Gaussian embeddings and thus is robust to the semantic uncertainty in label tokens.

\subsection{Influence to the Number of Label Tokens}

To probe how the number of label tokens impacts the final results, we present the performances under different numbers of label tokens in Figure~\ref{fig:tokens}.
Reading from the figure, we have analysis as follows.
1) The increasing number of label tokens could bring performance gain, but can not always produce improvements.
Specifically, the results peak with 3 label tokens and decreases when the number of label tokens enlarges.
The reason  may be that one token is too limited to express the label semantics while  too many tokens could bring noisy semantics disturbing the task prediction. 
2) Notably, all our reported results surpass the performances of prompt-learning-based baselines without uncertainty modeling in Table~\ref{tab:small-performance} and Table~\ref{tab:large-performance}.
Specifically, it produces the worst model performances when only one label token is employed.
Even though, the uncertainty modeling makes our method robust enough to outperform the vanilla prompt-learning methods.
This phenomenon again shows the semantic uncertainty of label tokens in SEP and certifies the importance of exploring verbalizer-uncertainty. 

Noted that the prompt tokens in our method are respectively derived from each  argument, thus the number of prompt tokens is fixed  instead of a hyper-parameter.

\begin{table}[t]
	\centering
	\caption{Model performances comparison between REP and P2G with different types of pretrained language models.}
	\setlength\tabcolsep{13pt}{
		\begin{tabular}{lcccc}
			\toprule
			\multirow{2}{*}{\textbf{Method}} & \multicolumn{2}{c}{\textbf{Base Benchmark}} & \multicolumn{2}{c}{\textbf{Large Benchmark}} \\
			& Accuracy(\%) & $\Delta$(\%) & Accuracy(\%) & $\Delta$(\%) \\
			\toprule
			\textbf{Baseline} & 59.99 & - & 67.05  & - \\
			\textbf{P2G} (BERT-base) & 61.78 & +1.79 & 68.16 & +1.11  \\
			\textbf{P2G} (RoBERTa-base) & 60.11 & +0.12 & 67.99 & +0.94  \\
			\textbf{P2G} (BERT-large) & 62.95 & +2.96 & 70.54 &  +3.49 \\
			\textbf{P2G} (RoBERTa-large) & 62.61 & +2.62 & 68.74 & +1.69  \\
			\bottomrule
	\end{tabular}}
	\label{tab:plms}
\end{table}

\subsection{Effects of PLMs}
\label{sec:appendix-plms}

Since P2G probes knowledge from PLMs, we  investigate how different PLMs affect the final performances.
We compare the model performances upon different PLMs with the strongest baselines exploring extra knowledge, KGModel and MCPredictor, for two benchmarks. 
The experimental results are shown in Table~\ref{tab:plms}, and we have observations as follows.
1) P2G upon different PLMs all surpasses the baselines, which  confirms the effectiveness of our method to probe knowledge in PLMs.
Besides, this also reveals that our method is PLM-agnostic, which could be generalized and adapted to other masked language models.
2) As the large-scale BERT and RoBERTa both advance their own basic versions, we could see that the larger the PLMs, the better performances could be achieved. 
We attribute this to that the large models could provide much richer knowledge to enhance the semantics. 
3) For both large and base models, we notice that BERT models outperform the RoBERTa models.
This phenomenon is consistent with what previous works~\citep{Liu2021PTuningVP,Shin2020ElicitingKF} observed that BERT models show more remarkable performances than RoBERTa models on prompt-learning-based methods, and this is worthy of investigation in the future.

\begin{table}[h]
	\centering
	\caption{
		Error analysis, where P2G mistakenly predicts e$_{c_3}$ as the subsequent event, while the correct one is e$_{c_4}$. 
	}
	\setlength\tabcolsep{6pt}
	\begin{tabular}{ccccc}
		\hline
		Events & Subject & Verb & Object & Indirect-O \\
		\hline
		e$_1$ & Clinton & broke & ground & NULL \\
		e$_2$ & Clinton & unveil & call & NULL \\
		e$_3$ & \textcolor{blue}{Clinton} &  \textcolor{blue}{boasted} & public & NULL\\
		e$_4$ & Clinton & call on & education & NULL \\
		e$_5$ & Clinton & increases & aid & NULL \\
		e$_6$ & Clinton & improves & education & NULL \\
		e$_7$ & Clinton & releases & program & inspiration \\
		e$_8$ & Clinton & hopes & education & better \\
		\hline
		\hline
		e$_{c_1}$ &  Clinton & broke & record & NULL      \\
		e$_{c_2}$ & Clinton & unveil & glue & NULL      \\
		\textcolor{blue}{e$_{c_3}$} &  \textcolor{blue}{Clinton} & \textcolor{blue}{boasted} & remark & NULL      \\
		\textcolor{red}{e$_{c_4}$}   &  \textcolor{red}{Students} & show & Clinton & supports      \\
		e$_{c_5}$ &  Clinton & increase & jet & NULL     \\
		\hline
	\end{tabular}
	\label{tab:complete-error}
\end{table}

\subsection{Error Analysis}

Although our method outperforms all baseline models, we still conduct an error analysis to probe its potential weaknesses.
We randomly chose 200 wrongly-predicted cases of P2G from the base benchmark, and present a typical one in Table~\ref{tab:complete-error} for analysis.
In this script instance, the given event chain tells events about Clinton's measures towards education, where the correct subsequent candidate, e$_{c_4}$, reflects students' attitudes towards Clinton's behaviors.
However, our model wrongly predicts e$_{3}$ as the subsequent event.
We owe such a wrong prediction to that P2G is strange to the newly emerging subject (``Students'') which has not appeared in the  event chain.
In this situation, the prompt computed from arguments cannot fully exploit the semantic relation between ``Students'' and the event chain.
On the contrary, the candidate event e$_{c_3}$ holds the same subject and verb as  e$_{3}$ in the given event chain, where the stronger semantic connections between e$_{c_3}$ and the given event chain could be correlated.
Hence, we infer that P2G may fall short in building semantic associations with the candidate event which holds different protagonists.
We think that this phenomenon points out an interesting direction to improve P2G better and inspire future works.

\section{Conclusion}

In this paper, we propose a novel prompt-learning-based method, Prompt2Gaussia (P2G), for Script Event Prediction (SEP).
To tackle the issues of prompt-uncertainty and verbalizer-uncertainty in SEP, we model the prompt and label tokens as Gaussian embeddings, where a prompt estimator and verbalizer estimator serve to estimate the probabilistic representations instead of the deterministic representations.
Experiments on two public benchmarks show the superiority of our method.
To our best knowledge, we are the first to explore prompt-learning for SEP.
In the future, we would like to adapt our method into other related tasks, such as event correlation reasoning.









\printcredits

\bibliographystyle{cas-model2-names}
\bibliography{sample-base}



\end{document}